\pdfoutput=1
\documentclass[10pt,twocolumn,letterpaper]{article}
\PassOptionsToPackage{table}{xcolor}
\usepackage[pagenumbers]{cvpr}
\usepackage{times}
\usepackage{helvet}
\usepackage{courier}
\usepackage[utf8]{inputenc}
\usepackage[T1]{fontenc}
\usepackage{multirow}
\usepackage{tcolorbox}
\usepackage{algorithm}
\usepackage{algorithmic}
\usepackage{microtype}
\usepackage{placeins}
\usepackage[symbol]{footmisc}
\definecolor{cvprblue}{rgb}{0.21,0.49,0.74}
\usepackage[pagebackref,breaklinks,colorlinks,allcolors=cvprblue]{hyperref}

\hypersetup{
  pdftitle={UniCAR-RL: Seeing Better before Thinking Deeper in Visual Mathematics},
  pdfauthor={Yuzhe Li, Hao Yan, Hao Wang, Xingchen Liu, Ya-Qi Yu, Jihao Wu, Minghui Liao, Wei Chen, Yuliang Liu}
}
\title{UniCAR-RL: Seeing Better before Thinking Deeper in Visual Mathematics}

\author{
Yuzhe Li\textsuperscript{1 *} \quad
Hao Yan\textsuperscript{1 *} \quad
Hao Wang\textsuperscript{2} \quad
Xingchen Liu\textsuperscript{1} \quad
Ya-Qi Yu\textsuperscript{2} \quad
\\
Jihao Wu\textsuperscript{2} \quad
Minghui Liao\textsuperscript{2\ensuremath{\dagger}} \quad
Wei Chen\textsuperscript{1} \quad
Yuliang Liu\textsuperscript{1\ensuremath{\ddagger}}
\\
\textsuperscript{1}Huazhong University of Science and Technology,
\quad
\textsuperscript{2}Huawei Inc.\\
\texttt{\{yzli, ylliu\}@hust.edu.cn}\\
\url{https://github.com/yzli9/UniCAR-RL}
}

\begin{document}

\maketitle
\raggedbottom

\makeatletter
\begingroup
\renewcommand{\thefootnote}{}
\renewcommand{\@makefntext}[1]{\noindent #1}
\begin{NoHyper}
\footnotetext{
\hspace*{1.5em}\textsuperscript{*} Equal contribution.\\
\hspace*{1.5em}\textsuperscript{\ensuremath{\dagger}} Project Leader.\\
\hspace*{1.5em}\textsuperscript{\ensuremath{\ddagger}} Corresponding author.
}
\end{NoHyper}
\endgroup
\makeatother

\begin{abstract}
Multimodal Large Language Models (MLLMs) often struggle with complex mathematical visual reasoning primarily due to a lack of fine-grained perception, causing initial visual hallucinations to directly trigger cascading reasoning failures. In traditional end-to-end reinforcement learning (RL), sparse rewards fail to decouple perceptual hallucinations from logical missteps, hindering targeted perception optimization. Alternatively, fine-tuning with perception-enhanced CoT data incurs high costs and hallucinations. 
In this paper, we address these challenges by proposing \textbf{UniCAR-RL}, an RL framework for annotation-free perception optimization. By explicitly decoupling the optimization of perception and reasoning during the training process, it achieves isolation and optimization of both capabilities. 
Specifically, UniCAR-RL consists of three synergistic branches:  1) a Caption-RL branch that optimizes perception capabilities through verifier-guided reasoning validation;
2) a Reasoning-RL branch that performs logical reasoning based on a gold image description to halt cascading errors;
3) a QA-RL branch that retains native end-to-end alignment to ensure robust question-answering performance. Experiments show that UniCAR-RL substantially improves MLLMs' mathematical and visual reasoning using only raw short-answer data. Furthermore, it demonstrates strong generalization across diverse architectures and scales.
\end{abstract}

\section{Introduction}
\label{sec:intro}

\begin{figure*}[t]
    \centering
    \includegraphics[width=\linewidth]{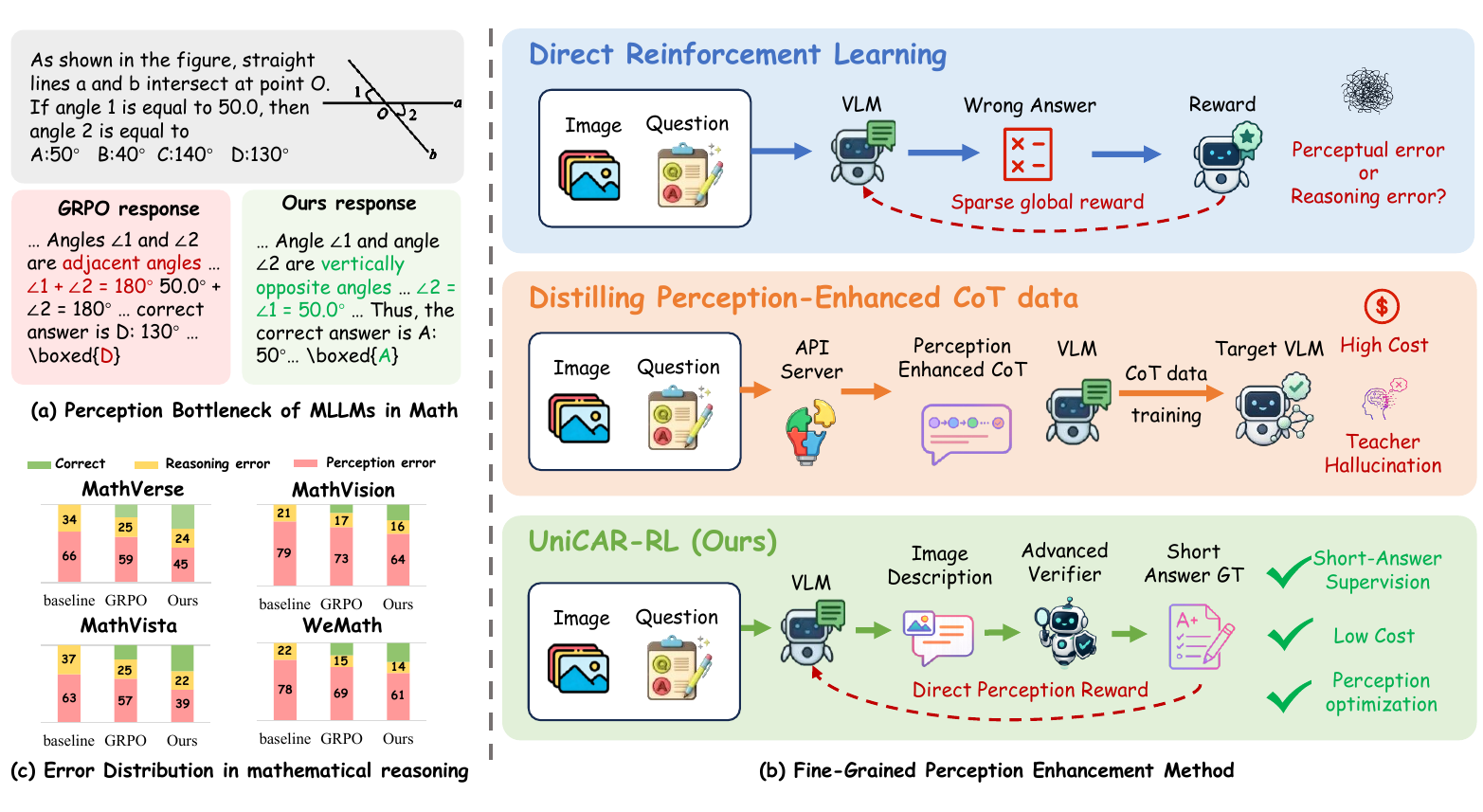}
    \caption{Motivation of proposed approach. (a) highlights the perception bottleneck of MLLMs in visual mathematical reasoning, but existing RL methods struggle to address it; (b) presents existing perception enhancement methods and their limitations; (c) provides a comparison of error distributions on visual mathematical reasoning tasks.}
    \label{fig:intro2}
\end{figure*}

Driven by the rapid advancements in Chain-of-Thought (CoT)~\cite{kojima2022large,wei2022chain} and Reinforcement Learning (RL)~\cite{shao2024deepseekmath,zhu2025shuffle} techniques, Multimodal Large Language Models (MLLMs) have achieved substantial progress in their reasoning capabilities for complex tasks~\cite{hu2025mplug, yan2026docseeker,li2024monkey}. However, the performance of existing MLLMs remains unsatisfactory when confronted with abstract visual reasoning tasks, such as math~\cite{mathverse,mathvision} and logic~\cite{xu2025visulogic, song2025visualpuzzles}. Extensive research~\cite{prco,yan2025visuriddles,visionsr1} shows that the primary bottleneck is not the models' insufficient reasoning skills, but rather the absence of a fine-grained perception capability. As illustrated in Fig.~\ref{fig:intro2}(a), when processing complex mathematical figures, models may easily misinterpret critical yet microscopic visual conditions, causing subsequent rigorous mathematical reasoning to be founded upon hallucinated visual premises. These initial visual misinterpretations fundamentally cause the breakdown of the mathematical reasoning.

As illustrated in Fig.~\ref{fig:intro2}(b), extensive recent research has emphasized the importance of enhancing fine-grained visual perception capabilities in MLLMs. Some recent w orks~\cite{guan2026codepercept,linger2025theorem} attempt to optimize models' perceptual abilities by distilling high-quality, perception-enhanced CoT data. However, these methods heavily rely on costly expert-level annotations and are inherently constrained by the hallucinations of teacher models. Alternatively, as an effective paradigm for boosting model performance on complex tasks, RL has been employed for optimization. However, sparse global rewards in RL fail to distinguish whether an incorrect prediction stems from visual misperception or logical errors, thereby rendering perception optimization highly inefficient.

To address the above challenges of existing methods, we propose Unified Caption-Answer-Reasoning Reinforcement Learning (\textbf{UniCAR-RL}), an RL framework for annotation-free perception optimization.  
As illustrated in Fig.~\ref{fig:intro2}(b), UniCAR-RL fundamentally disentangles perceptual evaluation from end-to-end black-box training by introducing an explicit image description process and an advanced verifier. 

Specifically, to obtain a pure perceptual feedback signal, the \textbf{Caption-RL Branch} of UniCAR-RL isolates the MLLM's internal reasoning module to generate an exclusive image description, and directly utilizes an advanced verifier's evaluation of this description as the reward. If the verifier successfully deduces the ground-truth answer relying solely on image description, it confirms the flawless extraction of visual information. Consequently, this verification mechanism returns an unambiguous direct perception reward to MLLMs, enabling precise optimization of perception capabilities while eliminating the interference of reasoning errors. 

Building upon the high-quality image descriptions,  \textbf{Reasoning-RL Branch} further enhances the model's logical reasoning capabilities. At this stage, this branch forces the model to perform independent mathematical reasoning based strictly on the gold caption verified in the preceding stage. This design of visual isolation fundamentally severs the error cascade of fine-grained perceptual hallucinations into the reasoning process, ensuring optimization signals are purely and intensely focused on enhancing logical deduction capabilities. 

Finally,  \textbf{QA-RL Branch} jointly optimizes perception and reasoning end-to-end, alongside maintaining the model's inherent question-answering proficiency. By synergizing these three branches, UniCAR-RL successfully achieves the joint optimization of perception and reasoning based on original short answers. The main advantages of UniCAR-RL are summarized as follows:

1) \textbf{Perception Enhancement via Raw Short-Answer Data.} As shown in Fig.~\ref{fig:intro2}(c), unlike traditional RL that primarily optimizes reasoning, this framework relies solely on raw short answers to simultaneously improve perceptual capabilities while preserving reasoning optimization.

2) \textbf{Performance Gains on Mathematical Tasks. } This framework achieves comprehensive gains across diverse mathematical benchmarks, delivering performance comparable to close-source proprietary models at a fraction of the training cost.

3) \textbf{Generalization Across Different Architectures and Sizes. } This framework achieves consistent and significant performance improvements across varying architectures and sizes. Furthermore, optimization efficacy exhibits limited sensitivity to the capabilities of the advanced verifier.

\section{Related Work}
\label{sec:related}

\paragraph{RL for Mathematical Reasoning.}

As reinforcement learning continually advances mathematical reasoning, GRPO~\cite{shao2024deepseekmath} and its variants~\cite{dapo,drgrpo,srpo} have become widely used post-training paradigms in this domain.
In multimodal mathematical reasoning, existing methods fall into cold-start RL and zero-RL. Cold-start RL bootstraps policies via curated reasoning traces or process supervision. We-Math 2.0~\cite{qiao2025we} structures training around mathematical knowledge, Vision-R1~\cite{visionr1} builds multimodal CoT data for GRPO initialization, and URSA~\cite{luo2026unlocking} combines a multimodal CoT foundation with process-supervised GRPO. Zero-RL directly optimizes MLLMs with rule-based rewards. MM-Eureka~\cite{mmeureka} scales rule-based RL on filtered image-text math data, while Shuffle-R1~\cite{zhu2025shuffle} improves RL efficiency through data-centric trajectory sampling and batch shuffling. R1-VL~\cite{zhang2025r1} introduces StepGRPO with dense step-wise rewards, and VL-Cogito~\cite{yuan2025vl} adopts curriculum RL with dynamic length rewards.

\paragraph{Visual Perception Enhancement in MLLMs.}
Fine-grained visual perception is essential for complex visual tasks, yet remains a major bottleneck for MLLMs~\cite{yan2025visuriddles, chen2025mathflow, linger2025theorem}. 
Existing methods mainly improve visual perception through perception-augmented trajectory training and perception-rewarded RL. The former typically distills high-quality perception-enhanced CoT data from advanced teacher models and then uses these data to enhance the perceptual capabilities of MLLMs. 
CodePercept~\cite{guan2026codepercept} and GeoCode~\cite{lin2026synthesizing} use executable programs to recover diagram structures, while MathFlow~\cite{chen2025mathflow} decouples visual extraction from downstream inference. ViRC~\cite{wang2025virc} and Geoint-R1~\cite{wei2025geoint} further structure the reasoning process with grounded reasoning chunks and auxiliary geometric constructions.
Perception-reward-guided RL methods~\cite{perceptionr1,cogflow,xing2025caprl} incorporate perceptual signals into reward design. They introduce rewards based on visual annotation consistency, visual knowledge internalization, and caption utility, respectively. Other studies reduce reliance on distilled perception data: VPPO~\cite{vppo} focuses policy updates on visually dependent tokens, Vision-SR1~\cite{visionsr1} self-verifies generated visual descriptions, PAPO~\cite{papo} derives grounding signals from visual perturbations, and NoisyRollout~\cite{liu2026noisyrollout} leverages noisy visual rollouts. Details of those works are provided in Appendix~\ref{appendix:relate_work}.

\begin{figure*}[!t]
    \centering
    \includegraphics[width=\linewidth]{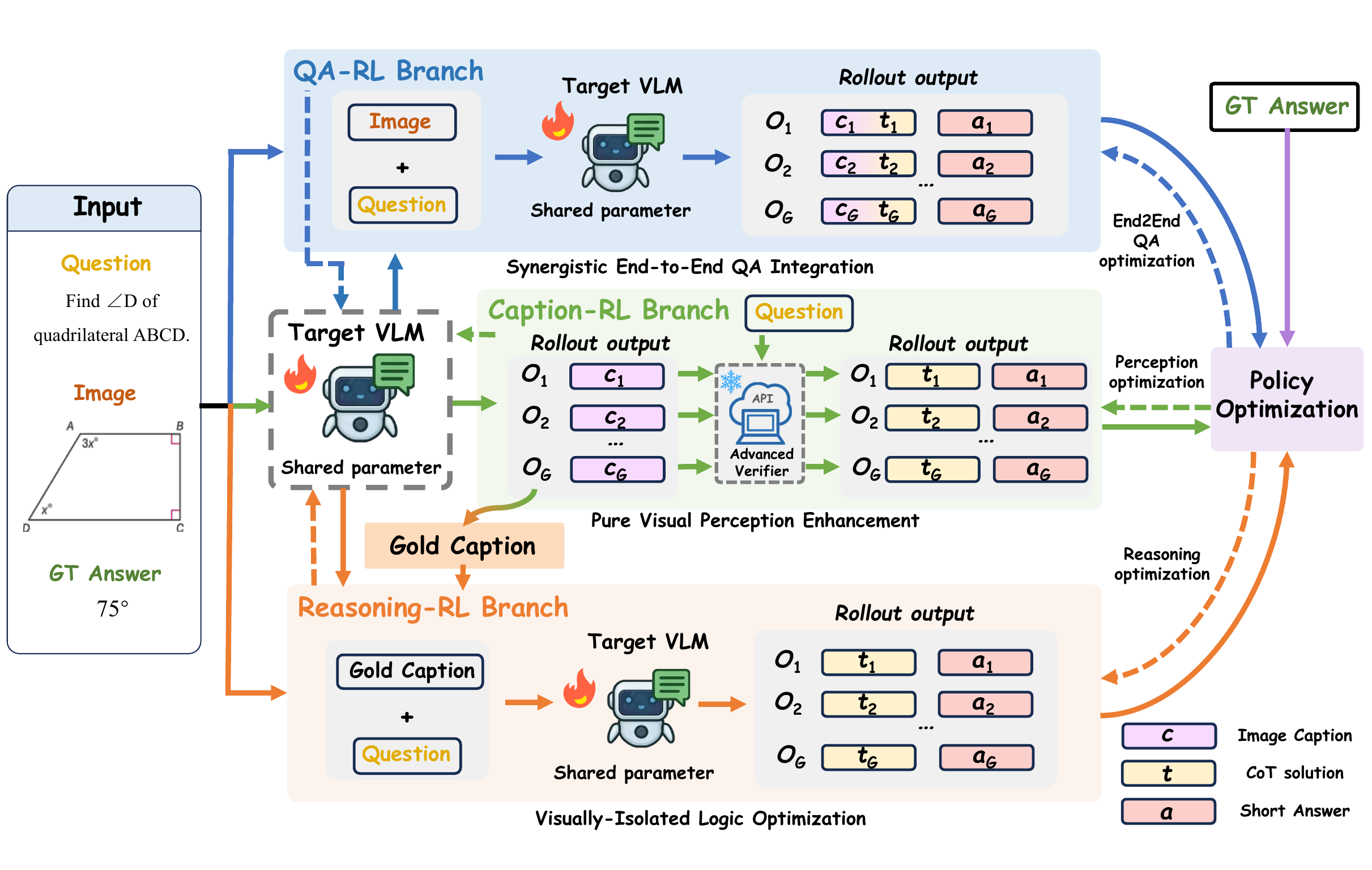}
    \caption{Overview of UniCAR-RL. It disentangles optimization into three branches: a) the \textbf{Caption-RL Branch} for pure visual perception enhancement, b) the \textbf{Reasoning-RL Branch} for visually-isolated logic optimization, and c) the \textbf{QA-RL Branch} for synergistic end-to-end QA integration.}
    \label{fig:pipeline}
\end{figure*}

\section{Method}
\label{sec:method}

\subsection{Overall Framework}
\label{sec:framework}

Visual mathematical reasoning demands two tightly coupled competencies: \textit{perception} faithfully extracting fine-grained visual details such as axis values, geometric marks, and symbolic annotations, and \textit{reasoning}  chains logical deductions toward a correct answer. A model can fail for either reason, yet conventional outcome-driven RL treats both failure modes identically: a wrong final answer yields a negative signal regardless of its root cause. Without a targeted signal, the perception bottleneck silently persists across training.

The key insight that motivates our approach is intuitive: \textit{the quality of a visual description can be objectively measured by whether a capable text-only solver can derive the correct answer from it alone.} This reformulation converts the inherently subjective problem of evaluating perception into a clean, annotation-free binary signal tied directly to task utility rather than surface-level textual fidelity.

Building on this insight, we propose \textbf{UniCAR-RL}, disentangling the optimization into three jointly updated branches. As illustrated in Fig.~\ref{fig:pipeline}, the \textbf{Caption-RL Branch} encourages broader, question-agnostic visual perception and evaluates generated descriptions by their task sufficiency. The \textbf{Reasoning-RL Branch} isolates logical reasoning in a purely textual setting, using the best available description as input. The \textbf{QA-RL Branch} then anchors both capabilities back to the native end-to-end task format. All three branches share the same policy parameters $\theta$ and are updated jointly in every iteration, so perception and reasoning co-evolve rather than being trained in isolated stages.

\subsection{Caption-RL Branch}
\label{sec:caprl}

\paragraph{Motivation.}
The Caption-RL Branch is built on a deliberate act of information withholding: the question $q$ is explicitly excluded from the model's input. A model that knows the question can selectively attend only to visually relevant elements; a model that does not must describe everything. By stripping $q$ from the context, we exert a natural pressure toward holistic, unbiased visual grounding.

\paragraph{Stage 1: Caption Generation.}
The policy $\pi_\theta$ generates $G$ caption rollouts from the image $I$ and a task prompt $p_{\rm cap}$, tasked with exhaustively extracting visual details without producing an answer:
\begin{equation}
    c_i \sim \pi_\theta(\cdot \mid I,\; p_{\rm cap}), \quad i=1,\dots,G.
\label{eq:caption_sampling}
\end{equation}
This design is inspired by the intuition behind EM-style optimization: the caption acts as a latent intermediate explanation, which is first inferred under the current policy and then used to optimize downstream reasoning. We do not perform EM in a strict probabilistic sense, since \(c^\star\) is chosen by reward-guided selection rather than posterior expectation, and the overall objective remains RL-based.

\paragraph{Stage 2: Verification.}
Each caption $c_i$ is evaluated by a \textit{proxy solver} $\mathcal{V}$, implemented as a text-only language model that accepts no visual input by design. Given only the caption $c_i$, the question $q$, and a task prompt $p_{\rm ver}$, $\mathcal{V}$ attempts to solve the problem entirely from the textual evidence at hand:
\begin{equation}
\hat{a}_i = \mathcal{V}(c_i,\; q,\; p_{\rm ver}), \quad i=1,\dots,G.
\label{eq:verifier_inference}
\end{equation}
Because $\mathcal{V}$ is a capable text-based reasoner with no access to $I$, the resulting reward serves as a practical proxy for caption sufficiency, though it may still contain noise from the verifier itself. The perception reward thus becomes:
\begin{equation}
r^{\rm cap}_i = \mathbb{I} \left[ \mathcal{D}(\mathrm{Extract}(\hat{a}_i), a^\star) \right], 
\label{eq:perception_reward}
\end{equation}
where $a^\star$ denotes the ground-truth answer. Missing a single critical visual element, such as a tick label, a right-angle mark, or a legend entry, will cause the text-only proxy solver to fail, sending a precise corrective signal back to the policy. 

\subsection{Reasoning-RL Branch}
\label{sec:rrl}

\paragraph{Gold Caption Selection.}
From the $G$ caption rollouts, we select the highest-reward caption as the selected \textit{gold caption} $c^\star$:
\begin{equation}
    c^\star = \arg\max_{c_i} \; r^{\rm cap}_i,
    \label{eq:gold_caption}
\end{equation}
breaking ties uniformly at random. Conditioning on $c^\star$ is deliberate: it provides a description that is simultaneously realistic (drawn from the policy's own distribution) and maximally informative.

\paragraph{Text-Only Reasoning.}
The image is withheld, and the model solves the problem from $c^\star$, $q$ and a task prompt $p_{\rm rsn}$:
\begin{equation}
    a'_j \sim \pi_\theta(\cdot \mid c^\star,\; q,\; p_{\rm rsn}), \quad j=1,\dots,G.
\end{equation}
This setup is structurally symmetric to the verification step in Caption-RL: there, the proxy solver $\mathcal{V}$ takes a caption and derives an answer; here, $\pi_\theta$ does the same but as the learner, so gradients flow directly through this rollout. In effect, the model is learning to be its own proxy solver. The reward follows the same rule-based extraction:
\begin{equation}
    r^{\rm rsn}_j=\mathbb{I}\!\left[\mathcal{D}(\mathrm{Extract}(a'_j),\;a^\star)\right].
\end{equation}
This branch also induces a \textit{natural curriculum}: as \textbf{Caption-RL Branch} improves, $c^\star$ becomes progressively richer, presenting \textbf{Reasoning-RL Branch} with increasingly demanding textual inputs. Stronger perception raises the ceiling for reasoning; stronger reasoning amplifies the value of accurate perception.

\subsection{QA-RL Branch}
\label{sec:qarl}

The two specialized branches target perception and reasoning separately. The \textbf{QA-RL Branch} serves as an integration stage: it trains the model to deploy both competencies under conditions matching downstream inference. The input image $I$, question $q$, and prompt $p_{\rm qa}$ mirror the test-time setting:
\begin{equation}
a_i \sim \pi_\theta(\cdot \mid I,\; q,\; p_{\rm qa}), \quad i=1,\dots,G,
\label{eq:qa_sampling}
\end{equation}
\begin{equation}
r^{\rm qa}_i = \mathbb{I} \left[ \mathcal{D}(\mathrm{Extract}(a_i), a^\star) \right].
\label{eq:qa_reward}
\end{equation}
By running this branch alongside the two specialized branches in every training iteration, the policy is continuously pulled back toward the task distribution it faces at deployment. The decoupled branches sharpen individual competencies; the QA branch ensures those competencies are always being woven back together.

\subsection{Unified GRPO Optimization}
\label{sec:optimization}

The three branches yield reward streams with heterogeneous semantics. To prevent any single branch from dominating the gradient, advantages are normalized independently within each rollout group. For branch $k \in \{\mathrm{cap}, \mathrm{rsn}, \mathrm{qa}\}$:
\begin{equation}
A_{i}^{(k)}
=
\frac{
r_{i}^{(k)}
-
\mathrm{mean}\!\left(\mathbf{r}^{(k)}\right)
}{
\mathrm{std}\!\left(\mathbf{r}^{(k)}\right)
}.
\label{eq:branch_advantage}
\end{equation}

Each branch's objective takes the standard clipped GRPO:
\begin{equation}
\begin{aligned}
J_{k}(\theta)
&=
\mathbb{E}\Bigg[
\frac{1}{G}\sum_{i=1}^{G}
\frac{1}{|o_i^{(k)}|}
\sum_{t=1}^{|o_i^{(k)}|}
\min\Big(
r_{i,t}^{(k)} A_i^{(k)},\\
&\quad \quad \quad \quad
\mathrm{clip}\big(
r_{i,t}^{(k)},\,1{-}\epsilon,\,1{+}\epsilon
\big) A_i^{(k)}
\Big)
\Bigg],
\end{aligned}
\label{eq:generic_objective}
\end{equation}

where the token-level importance ratio is
\begin{equation}
r_{i,t}^{(k)}
=
\frac{
\pi_\theta\!\left(
o_{i,t}^{(k)} \mid x^{(k)},o_{i,<t}^{(k)}
\right)
}{
\pi_{\rm old}\!\left(
o_{i,t}^{(k)} \mid x^{(k)},o_{i,<t}^{(k)}
\right)
}.
\label{eq:importance_ratio}
\end{equation}

All three objectives are optimized jointly via a single backward pass:
\begin{equation}
\theta \leftarrow \theta + \eta\,\nabla_\theta\!\Big(
J_{\rm cap}(\theta)
+
J_{\rm rsn}(\theta)
+
J_{\rm qa}(\theta)
\Big).
\label{eq:joint_update}
\end{equation}
This \emph{locally decoupled, globally co-evolving} paradigm is the defining characteristic of UniCAR-RL: each branch contributes a specialized gradient signal targeting a distinct competency, yet all signals are applied simultaneously to the same shared parameters within every single update step.

\section{Experiments}
\label{sec:exp}

\begin{table*}[!t]
\centering
\small
\setlength{\tabcolsep}{6pt}
\begin{tabular}{llccccc}
\toprule
Model & Venue &WeMath & MathVerse & MathVista & MathVision & Avg. \\
\midrule
\rowcolor{gray!10}
\multicolumn{7}{c}{\textit{Close-source Proprietary model}} \\
\midrule
GPT-4o~\cite{achiam2023gpt}
&  
& 68.8 & 50.2 & 63.8 & 30.4 & 53.4 \\

Claude-3.7-Sonnet~\cite{anthropic2025claude37sonnet}
&  
& 72.6 & 52.0 & 66.8 & 41.3 & 58.1 \\

\midrule
\rowcolor{gray!10}
\multicolumn{7}{c}{\textit{Models based on Qwen3-VL-Instruct}} \\
\midrule
Qwen3-VL-4B~\cite{qwen3vl}
& arXiv 2025
& 75.0 & 47.2 & 73.5 & 50.1 & 61.5 \\
Qwen3-VL-8B~\cite{qwen3vl}
& arXiv 2025
& \underline{79.4} & \underline{60.8} & 77.9 & 52.9 & \underline{67.8} \\
\midrule
UniCAR-RL-4B (Gemini 3 Pro)
& Ours
& 78.5 & 51.2 & \textbf{80.4} & \underline{53.3} & 65.9 \\
UniCAR-RL-8B (Gemini 3 Pro)
& Ours
& \textbf{82.5} & \textbf{63.4} & \underline{79.5} & \textbf{54.6} & \textbf{70.0} \\
\midrule

\rowcolor{gray!10}
\multicolumn{7}{c}{\textit{Models based on Qwen2.5-VL-Instruct-3B}} \\
\midrule

Qwen2.5-VL-3B~\cite{qwen25vl}
& arXiv 2025
& 55.9 & 33.6 & 59.4 & 22.5 & 42.9 \\

MMR1-3B-RL$^{\star}$~\cite{leng2025mmr1}
& CVPR-F 2026
& 64.5 & 35.3 & 65.3 & 23.5 & 47.2 \\

PAPO-D-3B$^{\star}$~\cite{papo}
& ICLR 2026
& 65.3 & 42.0 & 64.8 & 26.4 & 49.6 \\

Shuffle-R1-3B$^{\star}$~\cite{zhu2025shuffle}
& ICLR 2026
& 64.8 & 40.5 & 67.7 & 26.3 & 49.9 \\

\midrule
UniCAR-RL-3B (Qwen3.5-35B-A3B)
& Ours
& \underline{66.5} & \underline{41.8} & \underline{68.9} & \underline{27.3} & \underline{51.1} \\
UniCAR-RL-3B (Gemini 3 Pro)
& Ours
& \textbf{67.3} & \textbf{42.9} & \textbf{70.1} & \textbf{28.5} & \textbf{52.2} \\

\midrule
\rowcolor{gray!10}
\multicolumn{7}{c}{\textit{Models based on Qwen2.5-VL-Instruct-7B}} \\
\midrule
Qwen2.5-VL-7B~\cite{qwen25vl}
& arXiv 2025
& 61.7 & 42.5 & 67.3 & 25.7 & 48.3 \\

MMR1-7B-RL$^{\star}$~\cite{leng2025mmr1}
& CVPR-F 2026
& 71.5 & 46.0 & 70.8 & 29.3 & 54.4  \\

MM-Eureka-7B$^{\dagger}$~\cite{mmeureka}
& TMLR 2026
& 66.1 & 50.3 & 73.0 & 26.9 & 54.0 \\

NoisyRollout-7B$^{\dagger}$~\cite{liu2026noisyrollout}
& NeurIPS 2025
& 70.8 & 50.1 & 70.9 & 28.0 & 54.9 \\

R1-ShareVL-7B$^{\star}$~\cite{yao2026r1}
& NeurIPS 2025
& 69.8 & 52.8 & 75.1 & 29.5 & 56.8 \\

PAPO-D-7B$^{\star}$~\cite{papo}
& ICLR 2026
& 69.3 & 49.3 & 74.5 & 29.6 & 55.7 \\

Vision-R1-7B$^{\star}$~\cite{visionr1}
& ICLR 2026
& - & 50.4 & 71.2 & - & - \\

Vision-SR1-7B$^{\star}$~\cite{visionsr1}
& ICLR 2026
& 63.9 & 43.5 & 68.4 & 27.4 & 50.8 \\

Shuffle-R1-7B$^{\star}$~\cite{zhu2025shuffle}
& ICLR 2026
& 68.7 & 49.3 & 75.2 & 30.0 & 55.8 \\

Perception-R1-7B$^{\dagger}$~\cite{perceptionr1}
& ICLR 2026
& \underline{72.0} & \textbf{54.3} & 74.2 & 28.6 & 57.2 \\

\midrule

UniCAR-RL-7B (Qwen3.5-35B-A3B)
& Ours
& 71.9 & 53.0 & \underline{75.5} & \underline{30.2} & \underline{57.7} \\
UniCAR-RL-7B (Gemini 3 Pro)
& Ours
& \textbf{72.4} & \underline{53.7} & \textbf{76.0} & \textbf{30.6} & \textbf{58.2} \\

\bottomrule
\end{tabular}
\caption{Main results on visual mathematical reasoning benchmarks. The Avg. score is the arithmetic mean over WeMath, MathVerse, MathVista, and MathVision; CVPR-F denotes CVPR Findings. Models marked with $^{\star}$ are evaluated under our setting. Results marked with $^{\dagger}$ are sourced from previous work~\cite{perceptionr1}. The best value in each column is shown in \textbf{bold}, and the second-best is \underline{underlined}. For UniCAR-RL (\textit{Verifier}), the \textit{Verifier} denotes the advanced verifier used during training. This convention applies to all subsequent tables.}
\label{tab:math_results}
\end{table*}

\subsection{Experimental Setup}

\noindent\textbf{Dataset.} We utilize 30k samples from MMRL30k~\cite{zhu2025shuffle} during the training. In the evaluation phase, we conduct extensive testing on four highly challenging multimodal mathematical reasoning benchmarks, including WeMath~\cite{wemath}, MathVerse~\cite{mathverse}, MathVista~\cite{mathvista}, and MathVision~\cite{mathvision}. Furthermore, we verify the model's general capabilities on ChartQA~\cite{chartqa} and HallusionBench~\cite{hallbench}.

\noindent\textbf{Implementation Details.} Our experiments use Qwen2.5-VL~\cite{qwen25vl} and Qwen3-VL~\cite{qwen3vl} as backbones for full-parameter fine-tuning. Models are optimized using AdamW~\cite{adamw} with a learning rate of $1\times10^{-6}$, a weight decay of $1\times10^{-2}$, a micro-batch size of 8 per device, and a rollout number of 16 per input. The text-only training verifiers are Qwen3.5-35B-A3B~\cite{qwen3.5} and Gemini 3 Pro. For all locally evaluated models and benchmarks, decoding uses temperature 0.0, top-$p$ 0.8, top-$k$ 20, and presence penalty 0.0. For fair comparison, all evaluated models receive only the image and question as input during inference. Gemini-2.5-Flash~\cite{gemini25} is used only as a benchmark answer judge when required by the evaluation protocol.

\subsection{Main Results}
Table~\ref{tab:math_results} compares the proposed model with prior methods across four mathematical visual reasoning benchmarks. We draw the following observations:

1) \textbf{Significant performance gains.} By alleviating inherent perception bottlenecks of MLLMs, UniCAR-RL consistently delivers significant improvements across all evaluated backbone and scale configurations on visual mathematical reasoning benchmarks. It further achieves state-of-the-art average performance among the compared open-source models at comparable scales and remains competitive with proprietary models.

2) \textbf{Generalization across diverse backbones.}  By inherently enhancing visual perception through decoupling, UniCAR-RL demonstrates robust generalization, achieving substantial performance gains across architectures and scales. Notably, baselines with smaller sizes and lower initial performance tend to yield more significant improvements.

3) \textbf{Effect of advanced verifier.} Experiments clearly reveal the relationship between advanced verifier capability and model performance. A stronger verifier provides more precise perception reward signals, thereby more accurately guiding the model's optimization and significantly elevating the upper bound of its reasoning performance.

\begin{table*}[!t]
\centering
\small
\setlength{\tabcolsep}{8pt}
\begin{tabular}{lccccc}
\toprule
Configuration & WeMath & MathVerse & MathVista & MathVision & Avg.\\
\midrule
\multicolumn{6}{l}{\textit{Base model ablations (Verifier: Qwen3.5-35B-A3B)}} \\
\midrule
Qwen2.5-VL-3B \textcolor{gray}{\textrm{(base)}}   & 55.9 & 33.6 & 59.4 & 22.5 & 42.9 \\
\quad + UniCAR-RL           & 66.5\,{\scriptsize\color{teal}(+10.6)} & 41.8\,{\scriptsize\color{teal}(+8.2)} & 68.9\,{\scriptsize\color{teal}(+9.5)} & 27.3\,{\scriptsize\color{teal}(+4.8)} & 51.1\,{\scriptsize\color{teal}(+8.2)} \\
Qwen3-VL-4B \textcolor{gray}{\textrm{(base)}}     & 75.0 & 47.2 & 73.5 & 50.1 & 61.5 \\
\rowcolor{gray!10}
\quad + UniCAR-RL  & \textbf{77.8}\,{\scriptsize\color{teal}(+2.8)} & \textbf{50.4}\,{\scriptsize\color{teal}(+3.2)} & \textbf{79.6}\,{\scriptsize\color{teal}(+6.1)} & \textbf{52.9}\,{\scriptsize\color{teal}(+2.8)} & \textbf{65.2}\,{\scriptsize\color{teal}(+3.7)} \\
\midrule
\multicolumn{6}{l}{\textit{Verifier ablations}} \\
\midrule
Qwen3-VL-4B \textcolor{gray}{\textrm{(base)}}       & 75.0 & 47.2 & 73.5 & 50.1 & 61.5 \\
w/ Qwen3.5-35B-A3B                                   & 77.8\,{\scriptsize\color{teal}(+2.8)} & 50.4\,{\scriptsize\color{teal}(+3.2)} & 79.6\,{\scriptsize\color{teal}(+6.1)} & 52.9\,{\scriptsize\color{teal}(+2.8)} & 65.2\,{\scriptsize\color{teal}(+3.7)} \\
\rowcolor{gray!10}
w/ Gemini 3 Pro                            & \textbf{78.5}\,{\scriptsize\color{teal}(+3.5)} & \textbf{51.2}\,{\scriptsize\color{teal}(+4.0)} & \textbf{80.4}\,{\scriptsize\color{teal}(+6.9)} & \textbf{53.3}\,{\scriptsize\color{teal}(+3.2)} & \textbf{65.9}\,{\scriptsize\color{teal}(+4.4)} \\
\midrule
\multicolumn{6}{l}{\textit{Parameter scale ablations (Verifier: Qwen3.5-35B-A3B)}} \\
\midrule
Qwen3-VL-4B \textcolor{gray}{\textrm{(base)}}       & 75.0 & 47.2 & 73.5 & 50.1 & 61.5 \\
\quad + UniCAR-RL                                     & 77.8\,{\scriptsize\color{teal}(+2.8)} & 50.4\,{\scriptsize\color{teal}(+3.2)} & \textbf{79.6}\,{\scriptsize\color{teal}(+6.1)} & 52.9\,{\scriptsize\color{teal}(+2.8)} & 65.2\,{\scriptsize\color{teal}(+3.7)} \\
\addlinespace[3pt]
Qwen3-VL-8B \textcolor{gray}{\textrm{(base)}}        & 79.4 & 60.8 & 77.9 & 52.9 & 67.8 \\
\rowcolor{gray!10}
\quad + UniCAR-RL                                     & \textbf{82.5}\,{\scriptsize\color{teal}(+3.1)} & \textbf{63.4}\,{\scriptsize\color{teal}(+2.6)} & 79.5\,{\scriptsize\color{teal}(+1.6)} & \textbf{54.6}\,{\scriptsize\color{teal}(+1.7)} & \textbf{70.0}\,{\scriptsize\color{teal}(+2.3)} \\
\bottomrule
\end{tabular}
\caption{Component ablations of UniCAR-RL. We adopt Qwen3-VL-4B (Qwen3.5-35B-A3B) as the default configuration. We then conduct ablations on the base model, parameter scale, and verifier selection. Rows marked \textcolor{gray}{(base)} denote the pretrained model without UniCAR-RL training and green values report absolute improvement over the corresponding base.}
\label{tab:ablation}
\end{table*}

\begin{table*}[htbp]
\centering
\small
\setlength{\tabcolsep}{6pt}
\begin{tabular}{lccccc}
\toprule
\multirow{2}{*}{Captioner} & \multicolumn{5}{c}{Verifier-Solve Accuracy (\%)~$\uparrow$}\\
\cmidrule(lr){2-6}
 & WeMath & MathVerse & MathVista & MathVision & Avg.\\
\midrule
Qwen3-VL-4B (zero-shot)        & 80.8 & 62.4 & 76.6 & 65.9 & 71.4 \\
Qwen3-VL-4B + GRPO             & 82.1\,{\scriptsize\color{teal}(+1.3)} & 63.8\,{\scriptsize\color{teal}(+1.4)} & 77.8\,{\scriptsize\color{teal}(+1.2)} & 67.5\,{\scriptsize\color{teal}(+1.6)} & 72.8\,{\scriptsize\color{teal}(+1.4)} \\
Qwen3-VL-4B + UniCAR-RL & \textbf{84.9}\,{\scriptsize\color{teal}(+4.1)} & \textbf{65.5}\,{\scriptsize\color{teal}(+3.1)} & \textbf{82.5}\,{\scriptsize\color{teal}(+5.9)} & \textbf{70.8}\,{\scriptsize\color{teal}(+4.9)} & \textbf{75.9}\,{\scriptsize\color{teal}(+4.5)} \\

\bottomrule
\end{tabular}
\caption{Evaluation results of visual perception. We adopt Gemini 3 Pro as the unified verifier.}
\label{tab:caption_quality}
\end{table*}

\begin{table}[htbp]
\centering
\small
\setlength{\tabcolsep}{5pt}
\begin{tabular}{@{}lcc@{}}
\toprule
Model & ChartQA & HallBench \\
\midrule

\rowcolor{gray!10}
\multicolumn{3}{c}{\textit{Models around 4B parameters}} \\
\midrule
Qwen3-VL-4B
& 83.0 & 73.7 \\
+UniCAR-RL(Qwen3.5-35B-A3B)
& 87.9\,{\scriptsize\color{teal}(+4.9)} & 75.0\,{\scriptsize\color{teal}(+1.3)} \\
+UniCAR-RL(Gemini 3 Pro)
& 89.5\,{\scriptsize\color{teal}(+6.5)} & \textbf{75.5}\,{\scriptsize\color{teal}(+1.8)} \\

\midrule
\rowcolor{gray!10}
\multicolumn{3}{c}{\textit{Models around 8B parameters}} \\
\midrule
Qwen3-VL-8B
& 88.4 & 75.0 \\
+UniCAR-RL(Qwen3.5-35B-A3B)
& 90.2\,{\scriptsize\color{teal}(+1.8)} & 75.2\,{\scriptsize\color{teal}(+0.2)} \\
+UniCAR-RL(Gemini 3 Pro)
& 90.6\,{\scriptsize\color{teal}(+2.2)} & \textbf{75.4}\,{\scriptsize\color{teal}(+0.4)} \\
\bottomrule
\end{tabular}
\caption{Performance comparison on general multimodal benchmarks. HallBench stands for HallusionBench.}
\label{tab:general_results}
\end{table}

\subsection{Visual Perception Evaluation}
To quantitatively verify the perceptual enhancements of UniCAR-RL, we conduct a visual perception evaluation experiment. Specifically, we utilize the model to generate image descriptions for mathematical images, which are subsequently fed into a unified advanced verifier for problem-solving. Under this setup, the verifier’s accuracy objectively reflects the perception capability of the model. Table~\ref{tab:caption_quality} demonstrates that while standard GRPO brings negligible improvements, UniCAR-RL secures significant breakthroughs in description quality.

\subsection{General Tasks Evaluation}
\label{sec:general}

Although primarily optimized on visual mathematical data, UniCAR-RL mitigates the perception bottlenecks of MLLMs, yielding fine-grained perceptual enhancements that broadly benefit general multimodal tasks. We therefore further evaluate its generalization on a broader set of general tasks, using the system prompts provided in Appendix~\ref{app:general-benchmark-prompt}. Table~\ref{tab:general_results} reports the results on ChartQA~\cite{chartqa} and HallusionBench~\cite{hallbench}, while additional results are provided in Appendix~\ref{app:additional-perception}. The results show consistent performance gains across different model scales, demonstrating that the fine-grained perception enhancements of UniCAR-RL transfer beyond visual mathematical reasoning to diverse visual tasks.

\begin{figure*}[!t]
    \centering
    \includegraphics[width=\linewidth]{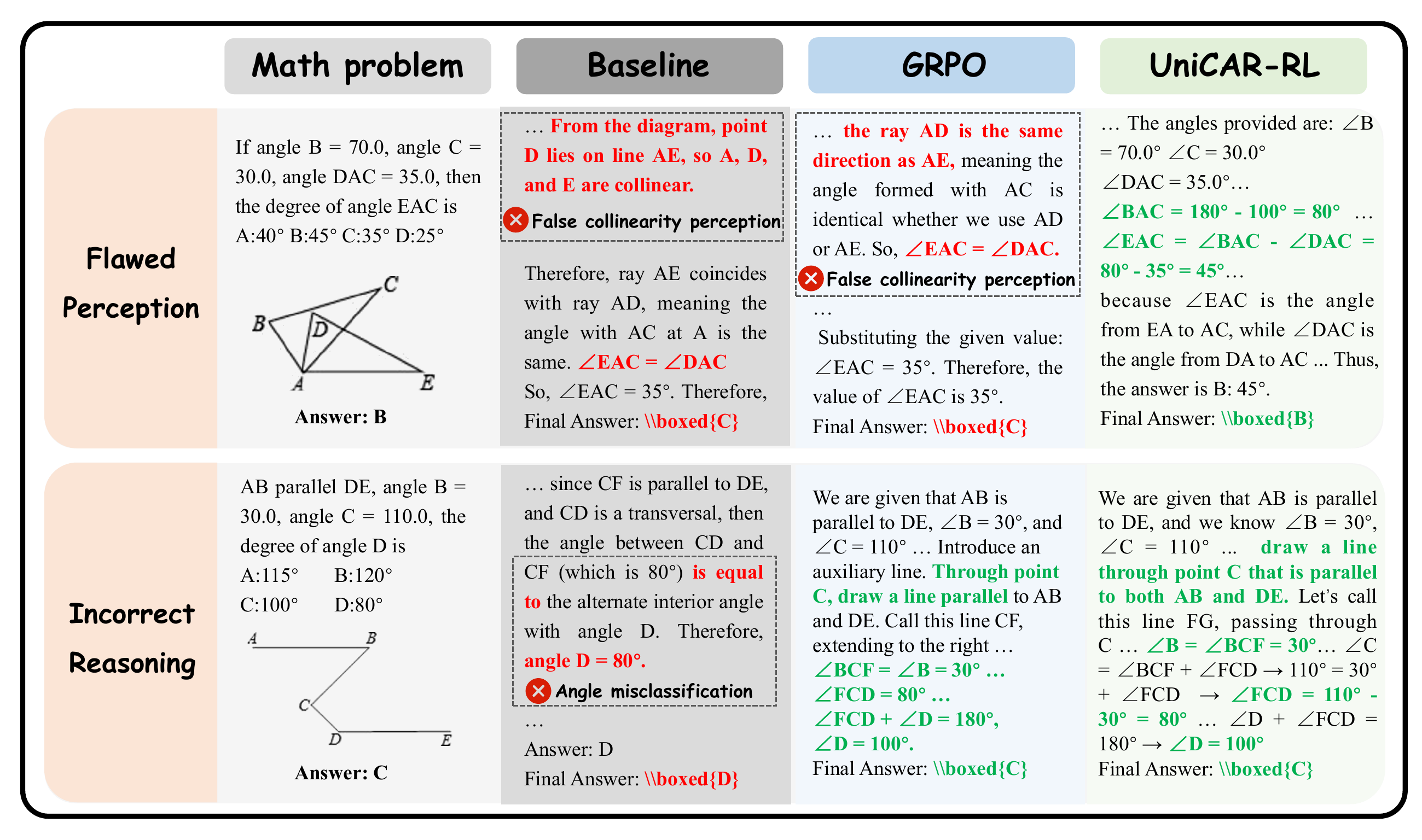}
    \caption{Qualitative comparison of mathematical reasoning tasks.}
    \label{fig:case}
\end{figure*}

\subsection{Ablation Study}
\label{sec:exp:ablation}

To investigate the contributions of each branch and component, we perform comprehensive ablation studies with Qwen3-VL-4B as the default setup.
\noindent\textbf{Ablation on Different Branches.} To verify the contribution of each branch, we progressively remove the \textbf{Reasoning-RL Branch}, \textbf{Caption-RL Branch}, and \textbf{QA-RL Branch}, gradually degrading UniCAR-RL back to the baseline. As demonstrated in Table~\ref{tab:branch}, the absence of any single branch leads to performance degradation, proving that all three branches play a crucial synergistic role in enhancing reasoning capabilities. More importantly, compared to the mere 1.3\% gain yielded by vanilla GRPO, the decoupled optimization of perception and reasoning significantly enhances the model's reasoning capabilities, achieving a 3.1\% improvement.

\noindent\textbf{Ablation on Key Components.} We further investigate the impact of different baselines, model sizes and verifiers on UniCAR-RL. As shown in Table~\ref{tab:ablation}, the results reveal three key observations. 
1) First, UniCAR-RL consistently yields significant performance enhancements across various base models. 2) Second, this positive trend remains highly stable across different parameter scales, verifying the strong scalability of our approach. 3) Third, the verifier serves purely as an auxiliary for problem-solving. Although a stronger verifier inherently leads to better performance, the marginal gains are relatively limited, suggesting that powerful open-source models can effectively substitute for proprietary ones in most scenarios.

\subsection{Qualitative Analysis}
To illustrate the advantages of UniCAR-RL, we present representative comparative examples in Fig.~\ref{fig:case}. In mathematical reasoning tasks, the failures of the baseline model primarily stem from visual perception errors and logical reasoning errors. Although standard RL methods, represented by GRPO, can help correct reasoning errors, they still fall short in optimizing the perception capabilities of MLLMs. In contrast, benefiting from the explicit decoupling of the perception and reasoning processes, UniCAR-RL can simultaneously and specifically optimize both visual perception and logical reasoning capabilities. Its synergistic multi-branch architecture not only retains the error-correction advantages of RL, but also precisely rectifies visual misjudgments through an independent branch, ultimately achieving substantial improvements in overall performance.

\begin{table*}[htbp]
\centering
\small
\setlength{\tabcolsep}{6pt}
\begin{tabular}{lccccc}
\toprule
Configuration & WeMath & MathVerse & MathVista & MathVision & Avg.\\
\midrule
\textbf{UniCAR-RL(full)}             & \textbf{78.5} & \textbf{51.2} & \textbf{80.4} & \textbf{53.3} & \textbf{65.9} \\
w/o Reasoning-RL Branch                     & 77.2 & 49.8 & 78.1 & 51.4 & 64.2 \\
w/o Caption-RL Branch (Vanilla GRPO)     & 76.3 & 48.5 & 76.0 & 50.8 & 62.8 \\
w/o QA-RL Branch (Base Qwen3-VL-4B)     & 75.0 & 47.2 & 73.5 & 50.1 & 61.5 \\
\bottomrule
\end{tabular}
\caption{Branch ablations of UniCAR-RL. By systematically removing each branch, UniCAR-RL progressively degrades to the standard baseline. The default setup uses Qwen3-VL-4B with the Gemini 3 Pro verifier.}
\label{tab:branch}
\end{table*}

\begin{table}[htbp]
\centering
\small
\setlength{\tabcolsep}{3pt}
\begin{tabular}{lcc}
\toprule
Method & Rollout & Total \\
\midrule
Vanilla GRPO (QA-RL) & 72.3 & 297.8 \\
Reasoning-RL & 68.4 & 292.7 \\
Caption-RL & 305.1 & 525.5 \\
\midrule
UniCAR-RL iteration (sum) & 445.8 & 1116.0 ($\sim$3.75$\times$) \\
\bottomrule
\end{tabular}
\caption{Training Cost Analysis of UniCAR-RL.}
\label{tab:time-cost}
\end{table}

\subsection{Training Efficiency Analysis}
\label{sec:training-cost}
To evaluate the training overhead associated with the performance gains of UniCAR-RL, we further analyze its training efficiency. We use Qwen2.5-VL-7B as the default backbone and measure both the rollout time and the total step time for each training branch. As shown in Table~\ref{tab:time-cost}, QA-RL branch and Reasoning-RL branch have training costs comparable to vanilla GRPO, while the additional overhead mainly comes from Caption-RL due to its extra verifier-based evaluation. A UniCAR-RL iteration takes approximately $3.75\times$ the time of a vanilla GRPO step. However, UniCAR-RL requires fewer optimization iterations in the observed training runs, with about 70 iterations compared with about 150 steps for vanilla GRPO, resulting in an overall training-time ratio of approximately $1.75\times$. Overall, UniCAR-RL achieves substantial performance gains with a moderate increase in training cost.

\section{Conclusion}
In this paper, we propose UniCAR-RL, a novel RL framework for annotation-free perception optimization that addresses the visual perception bottlenecks of MLLMs in complex visual mathematical reasoning. The core innovation lies in decoupling the optimization of perception and reasoning without relying on data distillation. By introducing a verifier to provide pure visual perception optimization signals and leveraging gold-description-based reasoning to deliver pure logical reasoning signals, UniCAR-RL can precisely co-optimize the model's fine-grained perception and logical reasoning capabilities. Extensive experiments show that our method exhibits robust generalization across diverse model architectures and sizes. Notably, because UniCAR-RL's performance gains stem from breaking the inherent perceptual bottlenecks of MLLMs, particularly in small sizes and earlier architectures, these enhancements naturally generalize to broad multimodal tasks.

\section*{Limitations}
Within the UniCAR-RL framework, the joint optimization of the synergistic Caption-RL branch, Reasoning-RL branch, and QA-RL branch inevitably increases the training cost, as discussed in Section~\ref{sec:training-cost}. Moreover, the framework is primarily validated on complex mathematical and a limited set of general visual tasks. Future work will explore its application to a broader range of multimodal scenarios.

\section*{Acknowledgments}
This research was supported by the NSFC 62576147.

{\small
\bibliographystyle{unsrt}
\bibliography{custom}

@String(CVPR   = {IEEE/CVF Conference on Computer Vision and Pattern Recognition})

@String(ICCV   = {IEEE/CVF International Conference on Computer Vision})

@String(ECCV   = {European Conference on Computer Vision})

@String(NIPS   = {Advances in Neural Information Processing Systems})

@String(ICLR   = {International Conference on Learning Representations})

@String(AAAI   = {AAAI Conference on Artificial Intelligence})

@String(ACL    = {Annual Meeting of the Association for Computational Linguistics})

@String(EMNLP  = {Conference on Empirical Methods in Natural Language Processing})

@String(FINDINGS = {Findings of the Association for Computational Linguistics})

@String(ICML   = {International Conference on Machine Learning})

@String(COLM   = {Conference on Language Modeling})

@String(CVPRF  = {IEEE/CVF Conference on Computer Vision and Pattern Recognition Findings})

@inproceedings{gemini25,
  title={{Gemini 2.5}: Pushing the frontier with advanced reasoning, multimodality, long context, and next generation agentic capabilities},
  author={Comanici, Gheorghe and others},
  booktitle={arXiv:2507.06261},
  year={2025}
}

@inproceedings{qwen25vl,
  title={{Qwen2.5-VL} technical report},
  author={Bai, Shuai and Chen, Keqin and Liu, Xuejing and Wang, Jialin and Ge, Wenbin and Song, Sibo and Dang, Kai and Wang, Peng and Wang, Shijie and Tang, Jun and others},
  booktitle={arXiv:2502.13923},
  year={2025}
}

@inproceedings{qwen3vl,
  title={{Qwen3-VL} technical report},
  author={Qwen Team},
  booktitle={arXiv:2511.21631},
  year={2025}
}

@inproceedings{mathvista,
  title={{MathVista}: Evaluating mathematical reasoning of foundation models in visual contexts},
  author={Lu, Pan and Bansal, Hritik and Xia, Tony and Liu, Jiacheng and Li, Chunyuan and Hajishirzi, Hannaneh and Cheng, Hao and Chang, Kai-Wei and Galley, Michel and Gao, Jianfeng},
  booktitle=ICLR,
  year={2024}
}

@inproceedings{mathverse,
  title={{MathVerse}: Does your multi-modal {LLM} truly see the diagrams in visual math problems?},
  author={Zhang, Renrui and Jiang, Dongzhi and Zhang, Yichi and Lin, Haokun and Guo, Ziyu and Qiu, Pengshuo and Zhou, Aojun and Lu, Pan and Chang, Kai-Wei and Qiao, Yu and Gao, Peng and others},
  booktitle=ECCV,
  pages={169--186},
  year={2024}
}

@inproceedings{mathvision,
  title={Measuring multimodal mathematical reasoning with {MATH-Vision} dataset},
  author={Wang, Ke and Pan, Junting and Shi, Weikang and Lu, Zimu and Zhan, Mingjie and Li, Hongsheng},
  booktitle=NIPS,
  volume={37},
  pages={95095--95169},
  year={2024}
}

@inproceedings{wemath,
  title={We-math: Does your large multimodal model achieve human-like mathematical reasoning?},
  author={Qiao, Runqi and Tan, Qiuna and Dong, Guanting and Wu, Minhui and Sun, Chong and Song, Xiaoshuai and Wang, Jiapeng and Gongque, Zhuoma and Lei, Shanglin and Zhang, Yifan and others},
  booktitle=ACL,
  pages={20023--20070},
  year={2025}
}

@inproceedings{chartqa,
  title={{ChartQA}: A benchmark for question answering about charts with visual and logical reasoning},
  author={Masry, Ahmed and Long, Do and Tan, Jia Qing and Joty, Shafiq and Hoque, Enamul},
  booktitle=FINDINGS,
  pages={2263--2279},
  year={2022}
}

@inproceedings{hallbench,
  title={{HallusionBench}: An advanced diagnostic suite for entangled language hallucination and visual illusion in large vision-language models},
  author={Guan, Tianrui and Liu, Fuxiao and Wu, Xiyang and Xian, Ruiqi and Li, Zongxia and Liu, Xiaoyu and Wang, Xijun and Chen, Lichang and Huang, Furong and Yacoob, Yaser and others},
  booktitle=CVPR,
  pages={14375--14385},
  year={2024}
}

@inproceedings{wei2022chain,
  title={Chain-of-thought prompting elicits reasoning in large language models},
  author={Wei, Jason and Wang, Xuezhi and Schuurmans, Dale and Bosma, Maarten and Ichter, Brian and Xia, Fei and Chi, Ed and Le, Quoc and Zhou, Denny},
  booktitle=NIPS,
  volume={35},
  pages={24824--24837},
  year={2022}
}

@inproceedings{shao2024deepseekmath,
  title={{DeepSeekMath}: Pushing the limits of mathematical reasoning in open language models},
  author={Shao, Zhihong and Wang, Peiyi and Zhu, Qihao and Xu, Runxin and Song, Junxiao and Bi, Xiao and Zhang, Haowei and Zhang, Mingchuan and Li, Y K and Wu, Y and Guo, Daya},
  booktitle={arXiv:2402.03300},
  year={2024}
}

@inproceedings{dapo,
  title={{DAPO}: An open-source {LLM} reinforcement learning system at scale},
  author={Yu, Qiying and Zhang, Zheng and Zhu, Ruofei and Yuan, Yufeng and Zuo, Xiaochen and Yue, Yu and Fan, Tiantian and Liu, Gaohong and Liu, Lingjun and Liu, Xin and others},
  booktitle=NIPS,
  volume={38},
  year={2025}
}

@inproceedings{vppo,
  title={Spotlight on Token Perception for Multimodal Reinforcement Learning},
  author={Huang, Siyuan and Qu, Xiaoye and Li, Yafu and Luo, Yun and He, Zefeng and Liu, Daizong and Cheng, Yu},
  booktitle=ICLR,
  year={2026}
}

@inproceedings{visionsr1,
  title={{Vision-SR1}: Self-rewarding vision-language model via reasoning decomposition and multi-reward policy optimization},
  author={Li, Zongxia and Yu, Wenhao and Liang, Zhenwen and Huang, Chengsong and Liu, Rui and Liu, Fuxiao and Chen, Jingxi and Yu, Dian and Boyd-Graber, Jordan and Mi, Haitao and Yu, Dong},
  booktitle=ICLR,
  year={2026}
}

@inproceedings{prco,
  title={Seeing with you: Perception-reasoning coevolution for multimodal reasoning},
  author={Miao, Ziqi and Jia, Haonan and Li, Lijun and Qian, Chen and Xiong, Yuan and Yan, Wenting and Shao, Jing},
  booktitle={arXiv:2603.28618},
  year={2026}
}

@inproceedings{visionr1,
  title={{Vision-R1}: Incentivizing reasoning capability in multimodal large language models},
  author={Huang, Wenxuan and Jia, Bohan and Cao, Shaosheng and Ye, Zheyu and Zhao, Fei and Xu, Zhe and Hu, Yao and Lin, Shaohui},
  booktitle=ICLR,
  year={2026}
}

@inproceedings{mmeureka,
  title={{MM-Eureka}: Exploring visual aha moment with rule-based large-scale reinforcement learning},
  author={Meng, Fanqing and Du, Lingxiao and Liu, Zongkai and Zhou, Zhixiang and Lu, Quanfeng and Fu, Daocheng and Shi, Botian and Wang, Wenhai and He, Junjun and Zhang, Kaipeng and Luo, Ping and Qiao, Yu and Zhang, Qiaosheng and Shao, Wenqi},
  booktitle={arXiv:2503.07365},
  year={2025}
}

@inproceedings{wu2024vstar,
  title={{V*}: Guided Visual Search as a Core Mechanism in Multimodal {LLMs}},
  author={Wu, Penghao and Xie, Saining},
  booktitle=CVPR,
  pages={13084--13094},
  year={2024}
}

@inproceedings{wang2025hrbench,
  title={Divide, Conquer and Combine: A Training-Free Framework for High-Resolution Image Perception in Multimodal Large Language Models},
  author={Wang, Wenbin and Ding, Liang and Zeng, Minyan and Zhou, Xiabin and Shen, Li and Luo, Yong and Yu, Wei and Tao, Dacheng},
  booktitle=AAAI,
  volume={39},
  number={8},
  pages={7907--7915},
  year={2025}
}

@misc{qwen3.5,
  title={{Qwen3.5}: Towards native multimodal agents},
  author={{Qwen Team}},
  url={https://qwen.ai/blog?id=qwen3.5},
  year={2026}
}

@inproceedings{kojima2022large,
  title={Large language models are zero-shot reasoners},
  author={Kojima, Takeshi and Gu, Shixiang Shane and Reid, Machel and Matsuo, Yutaka and Iwasawa, Yusuke},
  booktitle=NIPS,
  volume={35},
  pages={22199--22213},
  year={2022}
}

@inproceedings{zhu2025shuffle,
  title={{Shuffle-R1}: Efficient {RL} framework for multimodal large language models via data-centric dynamic shuffle},
  author={Zhu, Linghao and Guan, Yiran and Liang, Dingkang and Ju, Jianzhong and Luo, Zhenbo and Qin, Bin and Luan, Jian and Liu, Yuliang and Bai, Xiang},
  booktitle=ICLR,
  year={2026}
}

@inproceedings{yan2026docseeker,
  title={DocSeeker: Structured Visual Reasoning with Evidence Grounding for Long Document Understanding},
  author={Yan, Hao and Liu, Yuliang and Liu, Xingchen and Zhang, Yuyi and Liao, Minghui and Wu, Jihao and Chen, Wei and Bai, Xiang},
  booktitle=CVPR,
  pages={41140--41149},
  year={2026}
}

@inproceedings{hu2025mplug,
  title={m{PLUG}-{D}oc{O}wl2: High-resolution compressing for {OCR}-free multi-page document understanding},
  author={Hu, Anwen and Xu, Haiyang and Zhang, Liang and Ye, Jiabo and Yan, Ming and Zhang, Ji and Jin, Qin and Huang, Fei and Zhou, Jingren},
  booktitle=ACL,
  pages={5817--5834},
  year={2025}
}

@inproceedings{li2024monkey,
  title={Monkey: Image resolution and text label are important things for large multi-modal models},
  author={Li, Zhang and Yang, Biao and Liu, Qiang and Ma, Zhiyin and Zhang, Shuo and Yang, Jingxu and Sun, Yabo and Liu, Yuliang and Bai, Xiang},
  booktitle=CVPR,
  pages={26763--26773},
  year={2024}
}

@inproceedings{xu2025visulogic,
  title={{VisuLogic}: A benchmark for evaluating visual reasoning in multi-modal large language models},
  author={Xu, Weiye and Wang, Jiahao and Wang, Weiyun and Chen, Zhe and Zhou, Wengang and Yang, Aijun and Lu, Lewei and Li, Houqiang and Wang, Xiaohua and Zhu, Xizhou and Wang, Wenhai and Dai, Jifeng and Zhu, Jinguo},
  booktitle=ICLR,
  year={2026}
}

@inproceedings{song2025visualpuzzles,
  title={{VisualPuzzles}: Decoupling multimodal reasoning evaluation from domain knowledge},
  author={Song, Yueqi and Ou, Tianyue and Kong, Yibo and Li, Zecheng and Neubig, Graham and Yue, Xiang},
  booktitle=ICML,
  year={2026}
}

@inproceedings{yan2025visuriddles,
  title={{VisuRiddles}: Fine-grained perception is a primary bottleneck for multimodal large language models in abstract visual reasoning},
  author={Yan, Hao and Liu, Xingchen and Wang, Hao and Cao, Zhenbiao and Zheng, Handong and Yin, Liang and Su, Xinxing and Chen, Zihao and Wu, Jihao and Liao, Minghui and others},
  booktitle=ICLR,
  year={2026}
}

@inproceedings{linger2025theorem,
  title={Theorem-validated reverse chain-of-thought problem generation for geometric reasoning},
  author={Deng, Linger and Zhu, Linghao and Liu, Yuliang and Wang, Yu and Xie, Qunyi and Wu, Jingjing and Zhang, Gang and Zhu, Yingying and Bai, Xiang},
  booktitle=EMNLP,
  pages={718--735},
  year={2025}
}

@inproceedings{guan2026codepercept,
  title={{CodePercept}: Code-grounded visual {STEM} perception for {MLLMs}},
  author={Guan, Tongkun and Yang, Zhibo and Wan, Jianqiang and Yang, Mingkun and Guo, Zhengtao and Hu, Zijian and Luo, Ruilin and Chen, Ruizhe and Jiang, Songtao and Wang, Peng and Shen, Wei and Lin, Junyang and Yang, Xiaokang},
  booktitle=CVPR,
  year={2026}
}

@inproceedings{perceptionr1,
  title={{Perception-R1}: Advancing Multimodal Reasoning Capabilities of {MLLMs} via Visual Perception Reward},
  author={Xiao, Tong and Xu, Xin and Huang, Zhenya and Gao, Hongyu and Liu, Quan and Liu, Qi and Chen, Enhong},
  booktitle=ICLR,
  year={2026}
}

@inproceedings{cogflow,
  title={{CogFlow}: Bridging Perception and Reasoning through Knowledge Internalization for Visual Mathematical Problem Solving},
  author={Chen, Shuhang and Xu, Yunqiu and Xie, Junjie and Lu, Aojun and Feng, Tao and Huang, Zeying and Zhang, Ning and Sun, Yi and Yang, Yi and Yuan, Hangjie},
  booktitle=ICLR,
  year={2026}
}

@inproceedings{srpo,
  title={{SRPO}: Enhancing multimodal {LLM} reasoning via reflection-aware reinforcement learning},
  author={Wan, Zhongwei and Dou, Zhihao and Liu, Che and Zhang, Yu and Cui, Dongfei and Zhao, Qinjian and Shen, Hui and Xiong, Jing and Xin, Yi and Jiang, Yifan and others},
  booktitle=NIPS,
  volume={38},
  pages={153676--153713},
  year={2025}
}

@inproceedings{papo,
  title={{PAPO}: Perception-aware policy optimization for multimodal reasoning},
  author={Wang, Zhenhailong and Guo, Xuehang and Stoica, Sofia and Xu, Haiyang and Wang, Hongru and Ha, Hyeonjeong and Chen, Xiusi and Chen, Yangyi and Yan, Ming and Huang, Fei and Ji, Heng},
  booktitle=ICLR,
  year={2026}
}

@inproceedings{drgrpo,
  title={Understanding {R1}-zero-like training: A critical perspective},
  author={Liu, Zichen and Chen, Changyu and Li, Wenjun and Qi, Penghui and Pang, Tianyu and Du, Chao and Lee, Wee Sun and Lin, Min},
  booktitle=COLM,
  year={2025}
}

@inproceedings{leng2025mmr1,
  title={{MMR1}: Enhancing multimodal reasoning with variance-aware sampling},
  author={Leng, Sicong and Wang, Jing and Li, Jiaxi and Zhang, Hao and Hu, Zhiqiang and Zhang, Boqiang and Jiang, Yuming and Zhang, Hang and Li, Xin and Zhao, Deli and Lu, Wei and Rong, Yu and Sun, Aixin and Lu, Shijian},
  booktitle=CVPRF,
  pages={9075--9087},
  year={2026}
}

@inproceedings{chen2025mathflow,
  title={{MathFlow}: Enhancing the perceptual flow of {MLLMs} for visual mathematical problems},
  author={Chen, Shuhang and Yuan, Hangjie and Xu, Yunqiu and Liu, Pengwei and Feng, Tao and Cen, Jun and Huang, Zeying and Yang, Yi},
  booktitle=ACL,
  pages={967--992},
  year={2026}
}

@inproceedings{lin2026synthesizing,
  title={Synthesizing Multimodal Geometry Datasets from Scratch and Enabling Visual Alignment via Plotting Code},
  author={Lin, Haobo and Bai, Tianyi and Chen, Chen and Zhang, Jiajun and Zeng, Bohan and Zhang, Wentao and Yuan, Binhang},
  booktitle={arXiv:2602.18745},
  year={2026}
}

@inproceedings{wang2025virc,
  title={{ViRC}: Enhancing visual interleaved mathematical {CoT} with reason chunking},
  author={Wang, Lihong and Li, Liangqi and Feng, Weiwei and Wu, Jiamin and Miao, Changtao and Wu, Tieru and Ma, Rui and Zhang, Bo and Li, Zhe},
  booktitle=CVPR,
  pages={26144--26153},
  year={2026}
}

@inproceedings{wei2025geoint,
  title={{Geoint-R1}: Formalizing multimodal geometric reasoning with dynamic auxiliary constructions},
  author={Wei, Jingxuan and Jia, Caijun and Chen, Qi and He, Honghao and Sun, Linzhuang and He, Conghui and Wu, Lijun and Yu, Bihui and Tan, Cheng},
  booktitle=CVPR,
  pages={2547--2556},
  year={2026}
}

@inproceedings{xing2025caprl,
  title={{CapRL}: Stimulating dense image caption capabilities via reinforcement learning},
  author={Xing, Long and Dong, Xiaoyi and Zang, Yuhang and Cao, Yuhang and Liang, Jianze and Huang, Qidong and Wang, Jiaqi and Wu, Feng and Lin, Dahua},
  booktitle=ICLR,
  year={2026}
}

@inproceedings{zhang2025r1,
  title={{R1-VL}: Learning to Reason with Multimodal Large Language Models via Step-wise Group Relative Policy Optimization},
  author={Zhang, Jingyi and Huang, Jiaxing and Yao, Huanjin and Liu, Shunyu and Zhang, Xikun and Lu, Shijian and Tao, Dacheng},
  booktitle=ICCV,
  pages={1859--1869},
  year={2025}
}

@inproceedings{luo2026unlocking,
  title={Unlocking multimodal mathematical reasoning via process reward model},
  author={Luo, Ruilin and Zheng, Zhuofan and Wang, Lei and Wang, Yifan and Ni, Xinzhe and Lin, Zicheng and Jiang, Songtao and Yu, Yiyao and Shi, Chufan and Chu, Ruihang and others},
  booktitle=NIPS,
  volume={38},
  pages={49851--49899},
  year={2025}
}

@inproceedings{qiao2025we,
  title={{We-Math} 2.0: A versatile mathbook system for incentivizing visual mathematical reasoning},
  author={Qiao, Runqi and Yang, Peiqing and Wang, Yanzi and Wang, Xiaowan and Wan, Enhui and Dong, Guanting and Lang, Shiqiang and Zhou, Sitong and Xu, Yida and Zeng, Yuchen and others},
  booktitle=ICLR,
  pages={150171--150218},
  year={2026}
}

@inproceedings{yuan2025vl,
  title={{VL-Cogito}: Progressive curriculum reinforcement learning for advanced multimodal reasoning},
  author={Yuan, Ruifeng and Xiao, Chenghao and Leng, Sicong and Wang, Jianyu and Li, Long and Xu, Weiwen and Chan, Hou Pong and Zhao, Deli and Xu, Tingyang and Wei, Zhongyu and others},
  booktitle={arXiv:2507.22607},
  year={2025}
}

@inproceedings{liu2026noisyrollout,
  title={{NoisyRollout}: Reinforcing visual reasoning with data augmentation},
  author={Liu, Xiangyan and Ni, Jinjie and Wu, Zijian and Du, Chao and Dou, Longxu and Wang, Haonan and Pang, Tianyu and Shieh, Michael},
  booktitle=NIPS,
  volume={38},
  pages={3248--3282},
  year={2025}
}

@inproceedings{yao2026r1,
  title={{R1-ShareVL}: Incentivizing reasoning capabilities of multimodal large language models via {Share-GRPO}},
  author={Yao, Huanjin and Yin, Qixiang and Zhang, Jingyi and Yang, Min and Wang, Yibo and Wu, Wenhao and Su, Fei and Shen, Li and Qiu, Minghui and Tao, Dacheng and others},
  booktitle=NIPS,
  volume={38},
  pages={80424--80451},
  year={2025}
}

@inproceedings{achiam2023gpt,
  title={{GPT-4} technical report},
  author={Achiam, Josh and Adler, Steven and Agarwal, Sandhini and Ahmad, Lama and Akkaya, Ilge and Aleman, Florencia Leoni and Almeida, Diogo and Altenschmidt, Janko and Altman, Sam and Anadkat, Shyamal and others},
  booktitle={arXiv:2303.08774},
  year={2023}
}

@misc{anthropic2025claude37sonnet,
  title={{Claude 3.7 Sonnet} system card},
  author={{Anthropic}},
  note={System card},
  year={2025}
}

@inproceedings{adamw,
  title={Decoupled weight decay regularization},
  author={Loshchilov, Ilya and Hutter, Frank},
  booktitle=ICLR,
  year={2019}
}
}

\clearpage

\appendix
\onecolumn

\section{Visualization of the UniCAR-RL Training Workflow}
\label{appendix:visualization}
\refstepcounter{figure}
\label{fig:train_log}

\noindent
\begin{minipage}[t]{0.48\textwidth}
\fontsize{11.2pt}{13.7pt}\selectfont
\vspace{0pt}
To provide an intuitive view of UniCAR-RL, Fig.~\ref{fig:train_log} visualizes a training step across the three branches on a geometry problem. The \textbf{Caption-RL Branch} generates $G$ image descriptions and assigns per-rollout perception rewards via the advanced verifier, promoting the highest-scoring one as the gold caption. Since the verifier solves purely from text, its accuracy directly reflects faithfulness of visual evidence in the description. The \textbf{Reasoning-RL Branch} then reasons over this gold caption with the image withheld, while the \textbf{QA-RL Branch} performs standard end-to-end inference from the image and question. The three branches are executed sequentially within each iteration.

\vspace{0.6em}
\refstepcounter{section}
\label{sec:appendix_prompt}
\noindent\textbf{\thesection\quad Instruction Templates for UniCAR-RL}
\fontsize{11.2pt}{13.7pt}\selectfont
This section details the prompt templates used in UniCAR-RL training. We design three task-specific templates for the three synergistic branches of our framework, as illustrated in Fig.~\ref{fig:qa_prompt}, Fig.~\ref{fig:cap_prompt}, and Fig.~\ref{fig:rsn_prompt}. They correspond respectively to the \textbf{QA-RL Branch} for end-to-end answer alignment, the \textbf{Caption-RL Branch} for pure visual description extraction, and the \textbf{Reasoning-RL Branch} for text-only logical deduction.
\end{minipage}%
\hfill
\begin{minipage}[t]{0.48\textwidth}
\vspace{0pt}
\centering
\begin{tcolorbox}[colframe=black, colback=white, boxrule=0.5pt, arc=2pt, left=5pt, right=5pt, top=3pt, bottom=3pt]
\begin{center}\textbf{QA-RL Branch System Prompt}\end{center}
\texttt{You FIRST think about the reasoning process as an internal monologue and then provide the final answer. The reasoning process MUST BE enclosed within <think> </think> tags. The final answer MUST BE put in \textbackslash boxed\{\}.}
\end{tcolorbox}
\refstepcounter{figure}\label{fig:qa_prompt}
\raggedright{Figure~\ref{fig:qa_prompt}: System prompt used by the \textbf{QA-RL Branch}.}

\vspace{0.6em}
\centering
\begin{tcolorbox}[colframe=black, colback=white, boxrule=0.5pt, arc=2pt, left=5pt, right=5pt, top=3pt, bottom=3pt]
\begin{center}\textbf{Caption-RL Branch System Prompt}\end{center}
\texttt{As a helpful assistant, your role is to support visually impaired users by providing detailed and precise descriptions of input images. Your descriptions should be thorough, enabling users to understand and address any issues related to the images without needing to see them. Focus on offering visual aid without solving the questions for them.}
\end{tcolorbox}
\refstepcounter{figure}\label{fig:cap_prompt}
\raggedright{Figure~\ref{fig:cap_prompt}: System prompt used by the \textbf{Caption-RL Branch}.}
\end{minipage}

\vfill
\vspace{0.3em}
\begin{center}
\includegraphics[width=0.88\textwidth,height=0.44\textheight,keepaspectratio]{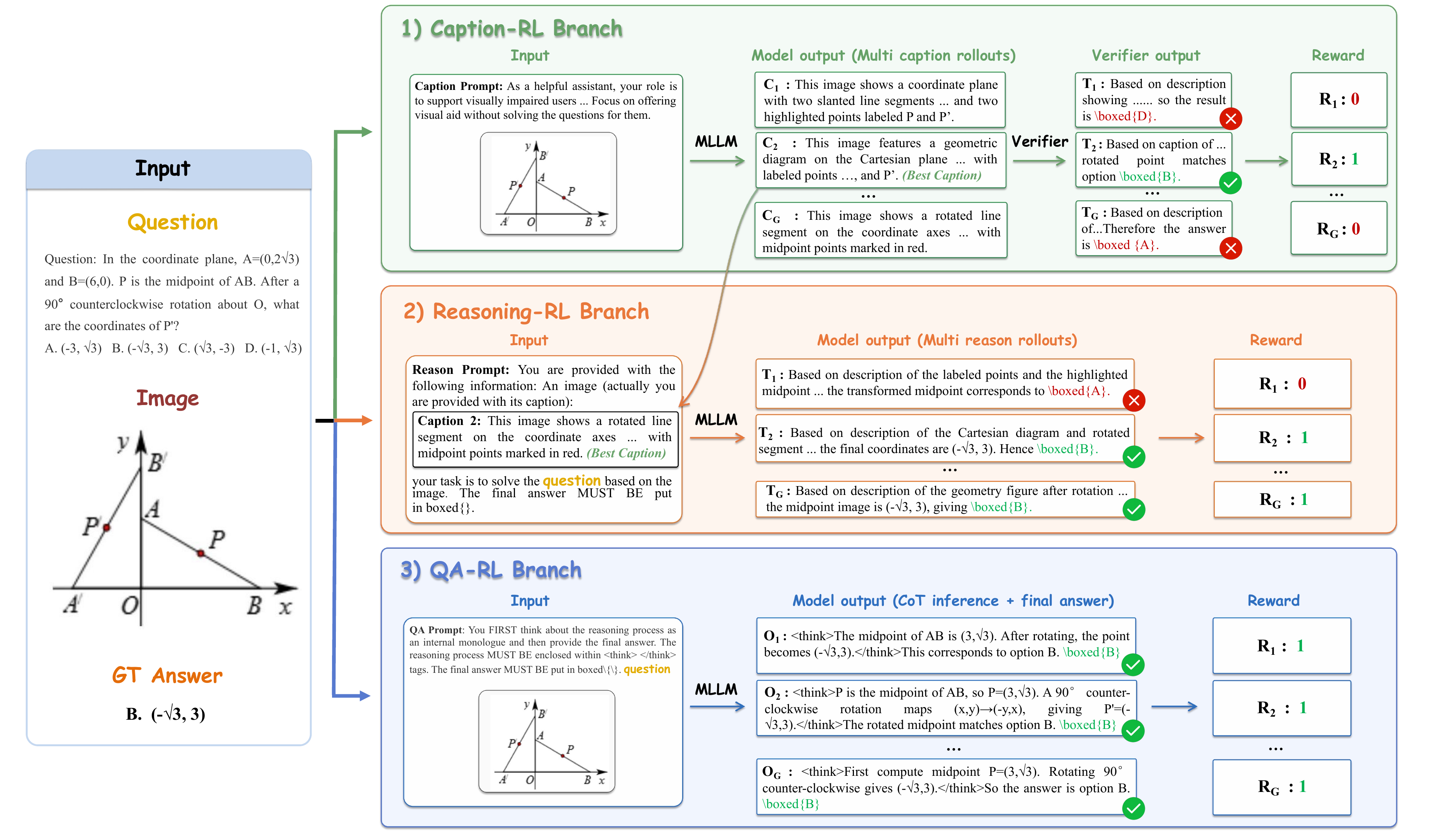}
\end{center}
\noindent{\textbf{Figure~\ref{fig:train_log}:} Visualization of a single UniCAR-RL training step. The three branches share the same policy network and are jointly optimized via a unified GRPO objective.}

\clearpage
\section{Related Work}
\label{appendix:relate_work}

\noindent
\begin{minipage}[t]{0.48\textwidth}
\fontsize{11.2pt}{13.7pt}\selectfont
\vspace{0pt}

To complement Section~\ref{sec:related}, Table~\ref{tab:math_reasoning_related_work} compares multimodal mathematical reasoning methods by \textbf{training paradigm} (Zero-RL vs.\ SFT+RL), \textbf{distillation} (whether teacher-generated CoT supervision is used, which may introduce additional cost and teacher hallucinations), and \textbf{perception} (whether visual perception is explicitly optimized). Existing methods in \textit{RL for Mathematical Reasoning} and \textit{Visual Perception Enhancement} typically focus on only part of these aspects. In contrast, UniCAR-RL combines Zero-RL, avoids distilled supervision, and explicitly optimizes visual perception within a unified framework.

\end{minipage}%
\hfill
\begin{minipage}[t]{0.48\textwidth}
\vspace{0pt}
\centering
\begin{tcolorbox}[colframe=black, colback=white, boxrule=0.5pt, arc=2pt, left=5pt, right=5pt, top=4pt, bottom=4pt]
\begin{center}\textbf{Reasoning-RL Branch System Prompt}\end{center}
\texttt{You are provided with the following information: an image (actually you are provided with its caption). Your task is to solve the question based on the image. The final answer MUST BE put in \textbackslash boxed\{\}.}
\end{tcolorbox}
\refstepcounter{figure}\label{fig:rsn_prompt}
\raggedright{Figure~\ref{fig:rsn_prompt}: System prompt used by the \textbf{Reasoning-RL Branch}, where the image is withheld and model reasons purely from caption.}
\end{minipage}

\vspace{10pt}

\begin{table}[!h]
\centering
\caption{Comparison of representative methods for multimodal mathematical reasoning. ``Train'' indicates the training paradigm, where ``Zero-RL'' denotes direct RL optimization from a pretrained or instruction-tuned model without task-specific cold-start SFT, ``SFT+RL'' denotes a cold-start supervised stage followed by reinforcement learning, and ``--'' indicates non-RL methods. ``Distill.'' indicates whether the method explicitly relies on distilled reasoning traces, teacher-generated supervision, or synthetic high-quality CoT/QA data. ``Percep.'' denotes whether the method explicitly enhances visual perception, grounding, diagram understanding, or visual reasoning capabilities.}
\label{tab:math_reasoning_related_work}
\scriptsize
\setlength{\tabcolsep}{5pt}
\renewcommand{\arraystretch}{1.2}
\resizebox{\textwidth}{!}{%
\begin{tabular}{l c l c c c c}
\toprule
\textbf{Method} & \textbf{Venue} & \textbf{Backbone} & \textbf{Size} & \textbf{Train} & \textbf{Distill.} & \textbf{Percep.} \\
\midrule
\multicolumn{7}{l}{\cellcolor{gray!10}\textit{Reinforcement Learning for Mathematical Reasoning}} \\
SRPO~\cite{srpo} & NeurIPS'25 & Qwen2.5-VL & 7B/32B & SFT+RL & $\checkmark$ & $\checkmark$ \\
Vision-R1~\cite{visionr1} & ICLR'26 & Qwen2.5-VL & 7B/32B/72B & SFT+RL & $\checkmark$ & $\checkmark$ \\
MM-Eureka~\cite{mmeureka} & TMLR'26 & Qwen2.5-VL & 7B/32B & Zero-RL & $\times$ & $\checkmark$ \\
R1-VL~\cite{zhang2025r1} & ICCV'25 & Qwen2-VL & 7B & SFT+RL & $\times$ & $\checkmark$ \\
URSA~\cite{luo2026unlocking} & NeurIPS'25 & Qwen2.5-Math & 8B & SFT+RL & $\checkmark$ & $\checkmark$ \\
We-Math 2.0~\cite{qiao2025we} & ICLR'26 & Qwen2.5-VL & 7B & SFT+RL & $\checkmark$ & $\checkmark$ \\
VL-Cogito~\cite{yuan2025vl} & arXiv'25 & Qwen2.5-VL & 7B & Zero-RL & $\times$ & $\checkmark$ \\
Shuffle-R1~\cite{zhu2025shuffle} & ICLR'26 & Qwen2.5-VL & 3B/7B & Zero-RL & $\times$ & $\checkmark$ \\
\midrule
\multicolumn{7}{l}{\cellcolor{gray!10}\textit{Visual Perception Enhancement in MLLMs}} \\
CodePercept~\cite{guan2026codepercept} & CVPR'26 & Qwen3-VL & 4B/8B/32B & SFT+RL & $\times$ & $\checkmark$ \\
GeoCode~\cite{lin2026synthesizing} & ICML'26 & Qwen2.5/3-VL & 7B & SFT & $\times$ & $\checkmark$ \\
MathFlow~\cite{chen2025mathflow} & ACL'26 & Qwen2-VL & 7B & SFT & $\checkmark$ & $\checkmark$ \\
ViRC~\cite{wang2025virc} & CVPR'26 & Qwen2.5-VL & 7B & SFT+RL & $\times$ & $\checkmark$ \\
Geoint-R1~\cite{wei2025geoint} & CVPR'26 & Qwen2.5-VL & 7B & SFT+RL & $\times$ & $\checkmark$ \\
Perception-R1~\cite{perceptionr1} & ICLR'26 & Qwen2/2.5-VL & 7B & Zero-RL & $\times$ & $\checkmark$ \\
CogFlow~\cite{cogflow} & ICLR'26 & Qwen2.5-VL & 7B & SFT+RL & $\times$ & $\checkmark$ \\
CapRL~\cite{xing2025caprl} & ICLR'26 & Qwen2.5/3-VL & 2B--8B & Zero-RL & $\times$ & $\checkmark$ \\
VPPO~\cite{vppo} & ICLR'26 & Qwen2.5-VL & 7B/32B & Zero-RL & $\times$ & $\checkmark$ \\
Vision-SR1~\cite{visionsr1} & ICLR'26 & Qwen2.5-VL & 7B & Zero-RL & $\times$ & $\checkmark$ \\
PAPO~\cite{papo} & ICLR'26 & Qwen2.5/3-VL & 2B/3B/7B & Zero-RL & $\times$ & $\checkmark$ \\
NoisyRollout~\cite{liu2026noisyrollout} & NeurIPS'25 & Qwen2.5-VL & 7B & Zero-RL & $\times$ & $\checkmark$ \\
\midrule
\rowcolor{blue!8}\textbf{UniCAR-RL (Ours)} & -- & Qwen2.5-VL, Qwen3-VL & 3B/4B/7B/8B & \textbf{Zero-RL} & $\times$ & $\checkmark$ \\
\bottomrule
\end{tabular}
}
\end{table}

\section{Additional Results on General Benchmarks}
\label{app:additional-perception}


To further evaluate the generalization of UniCAR-RL beyond visual mathematical reasoning, we extend the evaluation to a broader range of general visual tasks and model configurations. Beyond ChartQA and HallusionBench reported in Section~\ref{sec:general}, we further evaluate on V*Bench~\cite{wu2024vstar} and HR-Bench~\cite{wang2025hrbench}, covering complementary aspects of general visual understanding, including fine-grained visual search and high-resolution perception. As shown in Table~\ref{tab:additional-perception}, UniCAR-RL consistently improves performance across these additional tasks and architectures, further demonstrating its ability to generalize from visual mathematical reasoning to a broader range of general visual tasks.

\begin{table*}[t]
\centering

\caption{Additional perception results on ChartQA, HallBench, V*Bench and HR-Bench. Gemini 3 Pro is used as the verifier during training by default.}
\label{tab:additional-perception}
\begin{tabular}{lccccc}
\toprule
Model & ChartQA & HallBench & V*Bench & HR-Bench 4K & HR-Bench 8K \\

\midrule

Qwen3-VL-4B & 83.0 & 73.7 & 86.4 & 80.0 & 73.6 \\
+UniCAR-RL & 89.5 & 75.5 & 88.5 & 81.6 & 76.4 \\
Qwen3-VL-8B & 88.4 & 75.0 & 88.7 & 81.2 & 76.5 \\
+UniCAR-RL & 90.6 & 75.4 & 90.1 & 82.8 & 78.3\\
\bottomrule
\end{tabular}
\end{table*}

\section{Prompts for General Benchmarks}
\label{app:general-benchmark-prompt}

For reproducibility, we provide the prompts used for evaluating the general
multimodal benchmarks. All benchmarks are evaluated under the same chat
template, with the following shared instruction prepended to the
benchmark-specific question:

\begin{quote}
\small\ttfamily
You FIRST think about the reasoning process as an internal monologue and
then provide the final answer. The reasoning process MUST BE enclosed
within <think> </think> tags. The final answer MUST BE put in
\textbackslash boxed\{\}.
\end{quote}

The image is then provided together with the benchmark-specific question.
The corresponding question templates are described in
Fig.~\ref{fig:benchmark-prompt-templates}.

\newtcolorbox{promptbox}[2][5.0cm]{
    colback=white,
    colframe=black,
    coltitle=black,
    colbacktitle=white,
    boxrule=0.6pt,
    arc=2.5mm,
    outer arc=2.5mm,
    left=7pt,
    right=7pt,
    top=6pt,
    bottom=6pt,
    height=#1,
    valign=top,
    title=\textbf{#2},
    fonttitle=\small,
    titlerule=0.4pt,
    toptitle=3pt,
    bottomtitle=3pt,
    before skip=0pt,
    after skip=0pt,
}

\begin{figure*}[htbp]
\centering

\begin{minipage}[t]{0.485\textwidth}
\begin{promptbox}[2.7cm]{ChartQA}
\footnotesize
We directly use the original question without any
task-specific modification.

\vspace{6pt}
\ttfamily
\{question\}
\end{promptbox}
\end{minipage}
\hfill
\begin{minipage}[t]{0.485\textwidth}
\begin{promptbox}[2.7cm]{HallusionBench}
\footnotesize
We retain the original question and explicitly constrain
the final response to a binary answer.

\vspace{6pt}
\ttfamily
\{question\}\\
Your final answer can only be yes or no.
\end{promptbox}
\end{minipage}

\vspace{8pt}

\begin{minipage}[t]{0.485\textwidth}
\begin{promptbox}{V*Bench}
\footnotesize
Each question is formatted as a four-way
multiple-choice problem.

\vspace{6pt}
\ttfamily
Question: \{question\}\\[2pt]
Options:\\
A. \{option A\}\\
B. \{option B\}\\
C. \{option C\}\\
D. \{option D\}\\[2pt]
Please select the correct answer from the options above.
\end{promptbox}
\end{minipage}
\hfill
\begin{minipage}[t]{0.485\textwidth}
\begin{promptbox}{HR-Bench}
\footnotesize
HR-Bench-4K and HR-Bench-8K use the same
four-way multiple-choice prompt format.

\vspace{6pt}
\ttfamily
Question: \{question\}\\[2pt]
Options:\\
A. \{option A\}\\
B. \{option B\}\\
C. \{option C\}\\
D. \{option D\}\\[2pt]
Please select the correct answer from the options above.
\end{promptbox}
\end{minipage}

\vspace{7pt}

\caption{
\textbf{Task-specific prompt templates and output evaluation protocols.}
Placeholders enclosed in braces are replaced with the corresponding
benchmark questions and answer options at inference time.
}
\label{fig:benchmark-prompt-templates}
\end{figure*}

For output evaluation, ChartQA answers are extracted and evaluated using
the benchmark-specific answer matching procedure. For HallusionBench, we
extract the answer from the final boxed response and match it to
\texttt{yes}/\texttt{no}. For V*Bench and HR-Bench, the predicted option
letter is extracted and compared with the ground-truth choice.

\end{document}